\documentclass[10pt,twocolumn,letterpaper]{article}

\usepackage[pagenumbers]{cvpr} %

\usepackage{amsmath,amsfonts,bm}

\def\eqref#1{equation~\ref{#1}}

\def\1{\bm{1}}

\def\vc{{\bm{c}}}

\def\vm{{\bm{m}}}

\def\vp{{\bm{p}}}
\def\vq{{\bm{q}}}

\def\vs{{\bm{s}}}

\def\vv{{\bm{v}}}

\def\vx{{\bm{x}}}

\def\vz{{\bm{z}}}

\DeclareMathAlphabet{\mathsfit}{\encodingdefault}{\sfdefault}{m}{sl}
\SetMathAlphabet{\mathsfit}{bold}{\encodingdefault}{\sfdefault}{bx}{n}

\usepackage{dsfont}
\usepackage{wrapfig}
\usepackage{colortbl}
\usepackage{microtype}
\usepackage{subcaption}
\usepackage[dvipsnames]{xcolor}
\usepackage{comment}
\usepackage[accsupp]{axessibility}

\definecolor{baselinecolor}{gray}{.9}

\usepackage{pifont}
\newcommand{\cmark}{\ding{51}}%
\newcommand{\xmark}{\ding{55}}%

\usepackage{xspace}
\makeatletter
\DeclareRobustCommand\onedot{\futurelet\@let@token\@onedot}
\def\@onedot{\ifx\@let@token.\else.\null\fi\xspace}

\def\eg{\emph{e.g}\onedot}

\def\etc{\emph{etc}\onedot} \def\vs{\emph{vs}\onedot}

\makeatother

\usepackage{tocloft}

\makeatletter
\renewcommand{\numberline}[1]{%
  \@cftbsnum #1\@cftasnum\hspace*{1em}\@cftasnumb%
}
\makeatother

\newcommand{\tablestyle}[2]{\setlength{\tabcolsep}{#1}\renewcommand{\arraystretch}{#2}\centering\small}

\newcommand{\tablestylesmaller}[2]{\setlength{\tabcolsep}{#1}\renewcommand{\arraystretch}{#2}\centering\footnotesize}

\renewcommand{\paragraph}[1]{\vspace{.5em}\noindent\textbf{#1}}

\definecolor{cvprblue}{rgb}{0.21,0.49,0.74}
\usepackage[pagebackref,breaklinks,colorlinks,allcolors=cvprblue]{hyperref}

\def\paperID{11860} %
\def\confName{CVPR}
\def\confYear{2025}

\title{A Data-Centric Revisit of Pre-Trained Vision Models for Robot Learning}

\newcommand{\authorskip}{\hspace{2.5mm}}
\newcommand{\affiliationskip}{\hspace{2.5mm}}

\author{%
Xin Wen\textsuperscript{\mdseries1} \authorskip
Bingchen Zhao\textsuperscript{\mdseries2} \authorskip
Yilun Chen\textsuperscript{\mdseries3} \authorskip
Jiangmiao Pang\textsuperscript{\mdseries3} \authorskip
Xiaojuan Qi\textsuperscript{\mdseries1}\thanks{Corresponding author.} \vspace{0em}\\
\textsuperscript{\mdseries1}The University of Hong Kong \affiliationskip
\textsuperscript{\mdseries2}University of Edinburgh \affiliationskip
\textsuperscript{\mdseries3}Shanghai AI Laboratory \vspace{0em}\\
\small \tt \{wenxin, xjqi\}@eee.hku.hk
}

\begin{document}
\maketitle
\addtocontents{toc}{\protect\setcounter{tocdepth}{0}}
\begin{abstract}%
Pre-trained vision models (PVMs) are fundamental to modern robotics, yet their optimal configuration remains unclear. 
Through systematic evaluation, we find that while DINO and iBOT outperform MAE across visuomotor control and perception tasks, they struggle when trained on non-(single-)object-centric (NOC) data—a limitation strongly correlated with their diminished ability to learn object-centric representations.
This investigation indicates that the ability to form object-centric representations from the non-object-centric robotics dataset is the key to success for PVMs.
Motivated by this discovery, we designed SlotMIM, a method that induces object-centric representations by introducing a semantic bottleneck to reduce the number of prototypes to encourage the emergence of objectness as well as cross-view consistency regularization for encouraging multiview invariance.
Our experiments encompass pre-training on object-centric, scene-centric, web-crawled, and ego-centric data. Across all settings, our approach learns transferrable representations and achieves significant improvements over prior work in image recognition, scene understanding, and robot learning evaluations. When scaled up with million-scale datasets, our method also demonstrates superior data efficiency and scalability. 
Our code and models are publicly available at \url{https://github.com/CVMI-Lab/SlotMIM}.
\end{abstract}    
\begin{figure*}
\centering
\includegraphics[width=\textwidth]{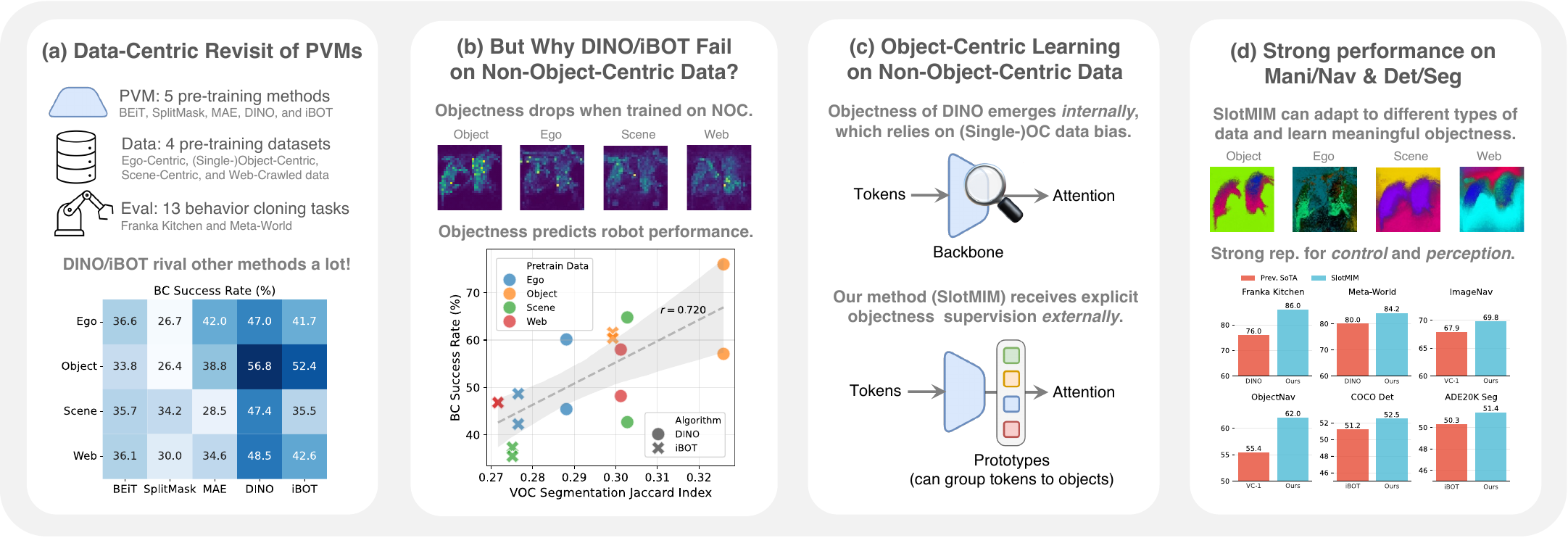}
\caption{\textbf{An overview of this paper.} (a) We conduct a comprehensive study evaluating pre-trained vision models (PVMs) on visuomotor control and perception tasks, analyzing how different pretraining (model, data) combinations affect performance. Our analysis reveals that DINO/iBOT excels while MAE underperforms.
(b) We investigate the performance drop of DINO/iBOT when trained on non-(single-)object-centric (NOC) data, discovering they struggle to learn objectness from NOC data---a capability that strongly correlates with robot manipulation performance.
(c) We introduce SlotMIM, which incorporates explicit objectness guidance during training to effectively learn object-centric representations from NOC data.
(d) Through scaled-up pre-training and evaluation across six tasks, we demonstrate that SlotMIM adaptively learns different types of objectness based on the pre-training dataset characteristics, outperforming existing methods.
} \label{fig:teaser}
\vspace{1em}
\end{figure*}

\section{Introduction}
Pre-trained vision models (PVMs) have become fundamental building blocks in modern computer vision and robotics. While these models have demonstrated remarkable success in traditional vision tasks, their optimal application in robot learning remains an open challenge. Recent works~\cite{radosavovic2023mvp,majumdar2023vc1} have shown promising results using masked autoencoders (MAEs)~\cite{he2021mae} pre-trained on ego-centric data (\eg, Ego4D~\cite{grauman2022ego4d}). However, as suggested by~\cite{dasari2023an}, learning on (single-)object-centric datasets like ImageNet~\cite{deng2009imagenet} can obtain better representations than on ego-centric data. Additionally, scene-centric data (\eg, COCO~\cite{lin2014microsoft} and Open Images~\cite{kuznetsova2020open}) are also relevant to robot contexts and more information-rich. Web-crawled data (\eg, CC12M~\cite{changpinyo2021cc12m}) offer another alternative, being easier to collect and more scalable. Towards scaling up PVMs for real-world robotic applications, it is important to explore diverse training data sources and revisit learning algorithms accordingly.
Our study is thus motivated by two questions: 1) Is MAE the optimal pre-training method for robot learning? 2) Is ego-centric data the best choice for visuomotor pre-training?

To systematically answer these questions, we construct a comprehensive benchmark that evaluates PVMs across both visuomotor control and perception tasks. For a fair comparison, we control the pre-training data scale at 241K images across different data sources and evaluate models on four diverse downstream tasks: two robot manipulation benchmarks (Franka Kitchen~\cite{gupta2019relay} and Meta-World~\cite{yu2019meta}), and two segmentation tasks (Pascal VOC~\cite{burns2024what} and ADE20K~\cite{zhou2017ade20k}). As shown in \cref{fig:rl_voc_ade_241k}, our first key finding challenges the role of MAE: DINO~\cite{caron2021dino} and iBOT~\cite{zhou2022ibot} significantly outperform MAE across all tasks, particularly when trained on object-centric data.
However, we observe a critical limitation: these models' performance degrades substantially when trained on scene-centric or non-(single-)object-centric (NOC) data. In coherence with the findings of~\cite{burns2024what}, we find that this performance drop strongly correlates with the models' diminishing ability to learn object-centric representations from NOC data. This suggests that the challenge lies not in the choice of pre-training method alone, but in maintaining object-centric learning capabilities when training on diverse data sources.

\begin{figure*}[t]
    \centering
    \includegraphics[width=.9\linewidth]{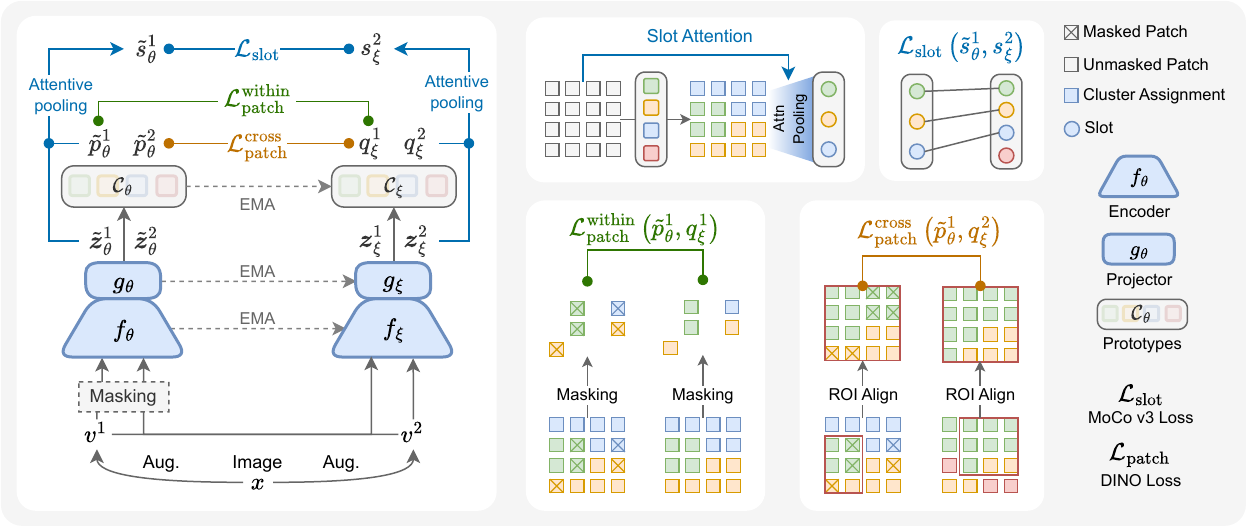}
    \caption{\textbf{Overview of SlotMIM.} Our framework extends iBOT by: 1) repurposing its within-view patch-level self-distillation for object discovery, 2) introducing a cross-view objective for semantic guidance, and 3) performing object-centric contrastive learning on slots (object features grouped from patches with matching cluster assignments). This approach provides explicit objectness supervision without requiring object-centric data, making it applicable to various types of NOC data (see \cref{fig:teaser}\textcolor{cvprblue}{c} for comparison and \cref{fig:teaser}\textcolor{cvprblue}{d} for results).
    }
    \label{fig:framework}
\end{figure*}

Motivated by these insights, we introduce SlotMIM (\cref{fig:framework}), a method that repurposes and integrates masked image modeling (MIM) and contrastive learning for effective representation learning from NOC datasets. The core idea of SlotMIM is to group patch-level image tokens into object-level feature abstractions, referred to as ``slots'', thereby decomposing NOC data into object-centric slots so that object-centric techniques can be effectively applied.
To make patch-level tokens more semantically aware for subsequent grouping, we enhance MIM with cross-view consistency regularization regarding prototype assignments. Additionally, we introduce a semantic bottleneck, which reduces the number of prototypes to encourage the emergence of semantic and objectness at patch-level token representations (see \cref{fig:compare_ibot_slotmim}).
Building on these semantically enriched patch tokens, we apply attentive pooling over the learned patch-level features, using prototypes to initialize object representations, thereby grouping patches into object-level slots. Contrastive learning~\cite{chen2021mocov3} is then applied to these slots to improve the discriminativeness of the learned representations.
Together, these designs enable us to perform effective representation learning from NOC data.

Our scaled experiments (\cref{fig:rl_ade_scale}) extend the analysis to million-scale datasets, revealing several surprising findings about the relationship between pre-training data and model performance. First, we observe an unexpected inverse scaling trend: while MAE's performance improves with more pre-training data, DINO and iBOT show degraded performance when scaled up. This indicates over-compression of representations: the learning objective reduces properties that are useful for manipulation tasks. Interestingly, we find that SlotMIM trained on ego-centric data learns to discover concepts adaptive to the pre-training data distribution, avoiding over-compression and consistently improving with more data. The resulting models—evaluated on six diverse tasks including Franka Kitchen, Meta-World, ObjectNav, ImageNav, COCO object detection and instance segmentation, and ADE20K semantic segmentation—demonstrate that SlotMIM consistently outperforms existing methods with better efficiency and scalability. Specifically, when pre-trained with just 241K samples, it already outperforms prior methods that used over 1M samples, including established approaches like MVP and VC-1. More significantly, while other methods saturate or degrade with increased NOC data, SlotMIM continues to improve and surpass previous methods that used 3$\times$ more data on both ADE20K and COCO benchmarks. This suggests that NOC data, when properly leveraged, can be a more scalable and efficient learning resource than previously thought, enabling new possibilities for scaling up PVMs.

In conclusion, our work challenges the prevailing trend of relying solely on scaling up ego-centric data with the MAE model to improve PVM transfer performance. Instead, we advocate for 1) finding a method (SlotMIM) that supports robust and scalable pre-training on different types of data, and 2) exploring multiple data types and using the one that best aligns with the target task (\eg, ego-centric data for manipulation tasks, scene-centric data for navigation tasks). This approach not only pushes the boundaries of what is possible with self-supervised learning but also aligns more closely with the practical needs of robot learning applications.

\begin{table*}[t]
\begin{minipage}{.57\textwidth}
\centering
\tablestyle{2.8pt}{1}
\begin{tabular}{lccccccc}
{Pre-train Data} & Source & ~ & {\#Image} & {\#Obj/Img} & {\#Class} & {Type} & {Video} \\
\toprule
{INet-241K} & ImageNet & ~ & {241K} & {1.7}  & 1000 & Object & \xmark  \\
{COCO+} & COCO & ~ & {241K} & {7.3} & 80 & Scene & \xmark \\
{CC-241K} & CC12M & ~ & {241K} & -- & -- & Web & \xmark \\
{Ego-241K} & Ego4D & ~ & {241K} & -- & -- & Ego & \cmark \vspace{1pt}\\
\toprule
\multicolumn{8}{l}{\footnotesize Object: Object-centric; Scene: Scene-centric; Web: Web-crawled; Ego: Ego-centric} \\
\end{tabular}
\end{minipage}\hfill
\begin{minipage}{0.42\textwidth}
\centering
\vspace{.9em}
\includegraphics[width=1\linewidth]{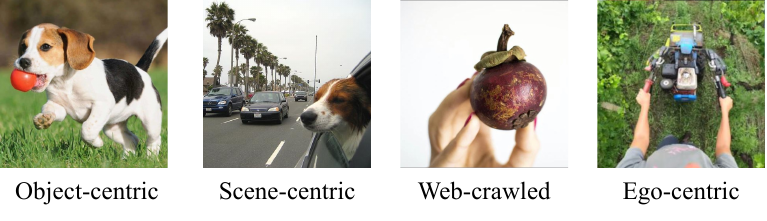}
\end{minipage}
\vspace{-.5em}
\caption{\textbf{Overview of pre-training datasets.} We uniformly sample subsets of 241K images from ImageNet, CC12M, and Ego4D. COCO+ is formed by combining \texttt{train} and \texttt{unlabeled} subsets of COCO. Ego4D frames are extracted at 0.2 fps and then sampled to subsets. 1.28M subsets are also considered in later experiments. For scene-centric data, we use the Open Images~\cite{kuznetsova2020open} dataset to scale up.}\label{tab:datasets}
\end{table*}

\section{Related Work}

\paragraph{Visual pre-training for robotics.}
Following the success of vision pre-training, robotics researchers have begun leveraging pre-trained vision models (PVMs) instead of training from scratch. As shown in~\cite{parisi2022unsurprising}, policies using PVMs can match or exceed the performance of those using ground-truth state information. A popular line of work is to train PVMs on ego-centric data (\eg, Ego4D~\cite{grauman2022ego4d}), including R3M~\cite{nairr3m}, MVP~\cite{xiao2022mvp,radosavovic2023mvp}, VIP~\cite{ma2023vip}, and VC-1~\cite{majumdar2023vc1}. However, a recent study~\cite{dasari2023an} shows that PVMs trained on ImageNet can also achieve competitive performance on downstream visuomotor tasks. Our work contributes to this line by expanding the scope of pre-training data to scene-centric and web-crawled data. More importantly, we also investigate methods beyond MAE~\cite{he2021mae}, which has been the predominant choice in many existing works~\cite{radosavovic2023mvp,dasari2023an,majumdar2023vc1,seo2023multi}. Other PVM paradigms include training from scratch with proper data augmentation~\cite{hansen2023pre}, from policy learning~\cite{ma2023vip}, from language~\cite{ma2023liv,vcond} or 3D~\cite{chen2024sugar}, \etc. Considering evaluation, while many works focus on manipulation tasks~\cite{hu2023pre,dasari2023an,nairr3m,ma2023vip,radosavovic2023mvp}, we also evaluate navigation tasks as in~\cite{majumdar2023vc1} and include perception-oriented tasks like segmentation and detection, thus providing a more complete picture.

\paragraph{Self-supervised visual representation learning.} 
Self-supervised representation learning aims at learning transferable features from unlabeled data~\cite{cmc,caron2018deepcluster,caron2020swav,caron2021dino,asano2020sela,chen2020simclr,bardes2022vicregl,assran2023self}. The field has converged on two main approaches: contrastive learning, which learns representations by comparing positive and negative examples~\cite{cmc,chen2020simclr,he2019moco}, and masked image modeling (MIM), which learns by reconstructing masked image regions~\cite{he2021mae,xie2022simmim}. While these methods have proven effective, their evaluation has largely focused on object-centric datasets like ImageNet-1K~\cite{deng2009imagenet}. Our work broadens this scope by studying self-supervised learning on large-scale non-(single-)object-centric datasets, and primarily evaluating the performance on robot learning tasks.

\paragraph{Learning on non-(single-)object centric (NOC) data.}
Several recent works have tackled self-supervised learning on non-object centric data~\cite{vangansbeke2021revisit,el2021large,xie2021pixpro,wang2021densecl,henaff2021detcon,henaff2022odin,wen2022slotcon,bai2022point}. Among them, the majority focuses on learning from scene-centric data and benefiting scene understanding tasks. The idea is primarily either to build pretext tasks on dense feature maps~\cite{wang2021densecl,xie2021pixpro,bai2022point}, object-centric groups~\cite{henaff2021detcon,henaff2022odin,wen2022slotcon}, or specialized data augmentation techniques~\cite{el2021large,vangansbeke2021revisit}. In addition, some works have also explored learning from uncurated datasets~\cite{caron2019noncurated,tian2021uncurated,bai2022point,oquab2023dinov2}, which mainly correspond to large-scale web-crawled datasets, and a main topic is data deduplication~\cite{oquab2023dinov2}. Our work contributes to this line of research by systematically studying the performance of self-supervised learning on multiple types of NOC data.

\section{When are PVMs More Effective for Visuomotor Control and Perception Tasks?}

\begin{figure*}[ht]
    \centering
    \includegraphics[width=\linewidth]{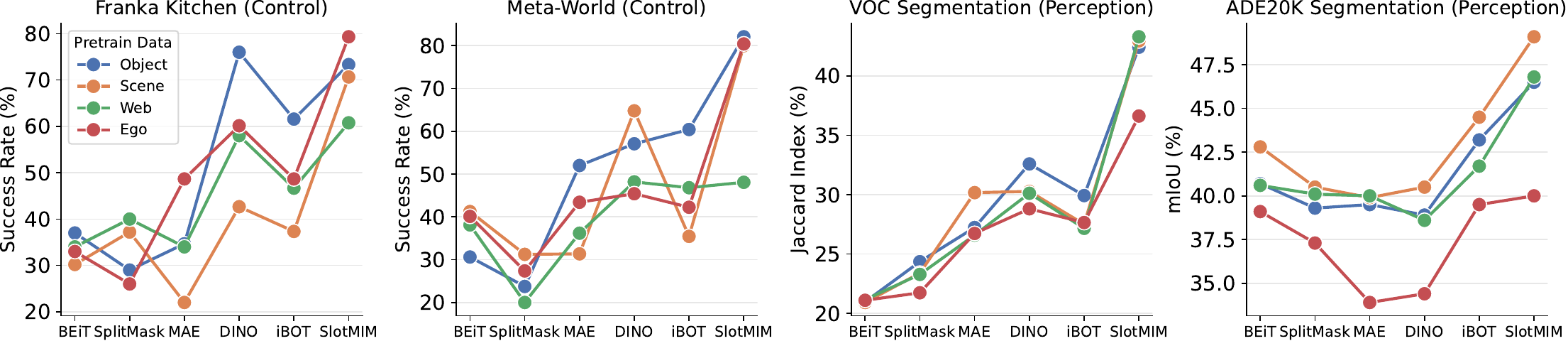}
    \caption{
    \textbf{Performance of PVMs trained with different (model, data) combinations on visuomotor control and perception tasks.} (241K scale, best viewed together with \cref{fig:teaser}\textcolor{cvprblue}{a})
    Our analysis of existing works reveals several key findings: 1) MAE with ego-centric data shows only moderate performance on visuomotor control tasks and performs poorly on ADE20K; 2) DINO and iBOT lead performance across all tasks, with their best models typically trained on object-centric data (except for ADE20K); 3) The top-3 models (DINO, iBOT, and MAE) struggle to learn effective representations for manipulation when trained on scene-centric data.
    Most notably, 4) SlotMIM (\cref{sec:method}) consistently outperforms prior methods regardless of whether it is pre-trained on object-centric data or not.
    }
    \label{fig:rl_voc_ade_241k}
\end{figure*}

\subsection{Pre-Training Setting}

\paragraph{Datasets.}
A common belief is that ego-centric data are the best for robot learning, primarily due to their contextual similarity to manipulation tasks~\cite{nairr3m,xiao2022mvp,ma2023vip,radosavovic2023mvp,majumdar2023vc1,vcond}. As suggested by~\cite{dasari2023an}, however, data diversity matters more for learning transferrable PVMs and traditional datasets like ImageNet~\cite{deng2009imagenet} can be more effective.
Our study considers the (single-)object-centric ImageNet~\cite{deng2009imagenet} and ego-centric Ego4D~\cite{grauman2022ego4d}. Moreover, our study also encompasses scene-centric data (\eg, COCO~\cite{lin2014microsoft}), which are also close to robot contexts and are more information-rich. We also consider web-crawled data (\eg, CC12M~\cite{changpinyo2021cc12m}), as they are easier to collect and more friendly to scale up. As shown in \cref{tab:datasets}, experiments in this section control the data scale to 241K uniformly sampled images from each dataset. This ensures fair comparison and efficient pre-training, allowing us to expand pre-training to more (method, dataset) combinations. %

\paragraph{Methods.}
Another widely-accepted belief is that MAE~\cite{he2021mae} is one of the best pre-training methods for PVMs~\cite{xiao2022mvp,radosavovic2023mvp,seo2023multi,majumdar2023vc1}, which remains unchallenged by~\cite{dasari2023an}. We are interested in whether expanding the search space to NOC data could reveal other strong candidates. While existing works have repeatedly demonstrated MAE's superiority over ResNets~\cite{nairr3m,ma2023vip} and robot-oriented ViTs~\cite{radford2021clip,chen2021mocov3}, we compare MAE with several other ViT pre-training methods: BEiT~\cite{bao2021beit}, SplitMask~\cite{el2021large}, DINO~\cite{caron2021dino}, and iBOT~\cite{zhou2022ibot}. These methods were selected based on their strong performance on perception tasks like detection and segmentation.
Given the significant computational cost of reproducing pre-training (including MAE) on multiple datasets, this selection provides a good representative set of methods.

\paragraph{Training.}
We use ViT-B/16~\cite{dosovitskiy2021vit} as the backbone. At 241K data scale, all methods are trained for 800 epochs. At 1.28M data scale, we train for 400 epochs. The optimization hyperparameters follow official settings of each method.

\begin{table}[t]
    \centering
    \tablestylesmaller{2pt}{1}
    \begin{tabular}{lcccccc}
    {Benchmark Suite} & RGB & Proprio. & {Physics} & {Action} & {Goal} & {Learning}\\
    \toprule
    Franka Kitchen~\cite{gupta2019relay} & \cmark & \xmark & \cmark & Continu. & -- & IL \\
    Meta-World~\cite{yu2019meta} & \cmark & \xmark & \cmark & Continu. & -- & IL \\
    ObjectNav~\cite{batra2020objectnav} & \cmark & \cmark & \xmark & Discrete & Class & IL \\
    ImageNav~\cite{zhu2017imagenav} & \cmark & \xmark & \xmark & Discrete & Image & RL \\
    \toprule
    \end{tabular}
    \vspace{1.5mm}
    \includegraphics[width=1\linewidth]{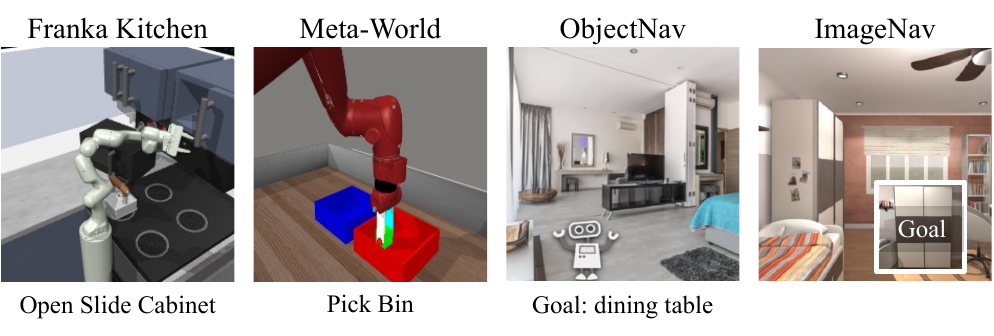}
    \vspace{-7mm}
    \caption{\textbf{Overview of robot manipulation and navigation tasks.} Bottom: example tasks of each benchmark suite.}\label{tab:robo_tasks}
\end{table}

\subsection{Evaluation Setting}
We expect a generalizable PVM to benefit both visuomotor control tasks (\eg, manipulating household objects and navigating indoor environments) and perceptual understanding tasks (\eg, recognizing objects and scenes, and correctly localizing them). While these tasks may share some common features, they can also require contrasting properties (\eg, grasping \vs navigation).
Therefore, we evaluate PVMs on as diverse tasks as possible to understand their properties and how they interact with (pre-training) dataset biases. An overview is shown in \cref{tab:robo_tasks}.
Due to computational constraints, this section focuses on manipulation (control) and segmentation (perception) tasks, with the strongest PVMs selected for navigation (control) and detection (perception) tasks in later sections.

\begin{figure}[t]
    \centering
    \includegraphics[width=.8\linewidth]{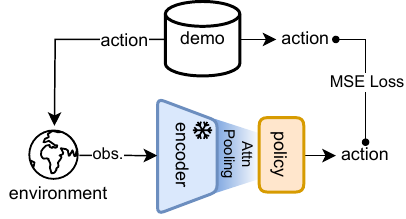}
    \caption{\textbf{Behavior cloning with attentive probing.} An additional token is trained with cross-attention (trainable) to gather information from all patch tokens from the backbone (frozen), and fed to the policy to learn from expert demonstrations via behavior cloning.}
    \label{fig:bc_loop}
\end{figure}

\begin{figure*}[ht]
    \centering
    \begin{subfigure}[b]{0.35\linewidth}
        \centering
        \includegraphics[width=\textwidth]{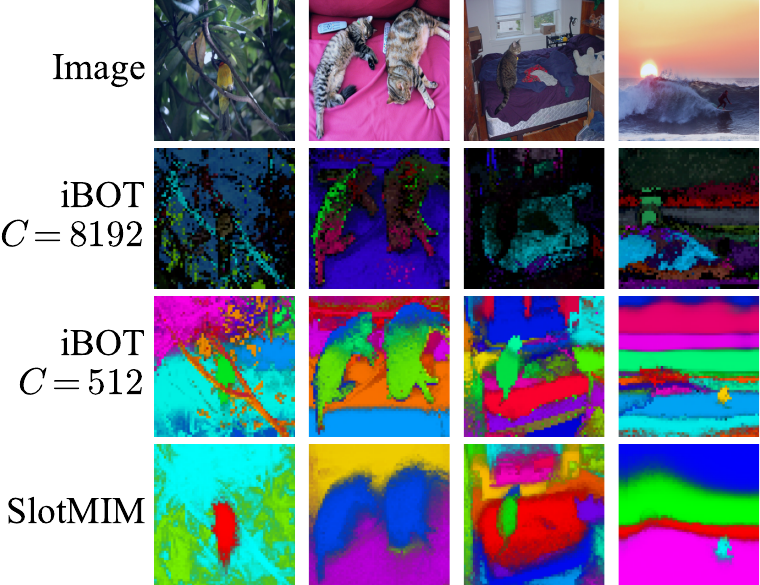}
        \caption{\textbf{Clustering assignment of patch tokens.} Each patch is assigned to its nearest-neighbor prototype, with different colors indicating different prototypes.}
        \label{fig:attns_ibot_slotmim}
    \end{subfigure}
    \hfill
    \begin{subfigure}[b]{0.597\linewidth}
        \centering
        \includegraphics[width=\textwidth]{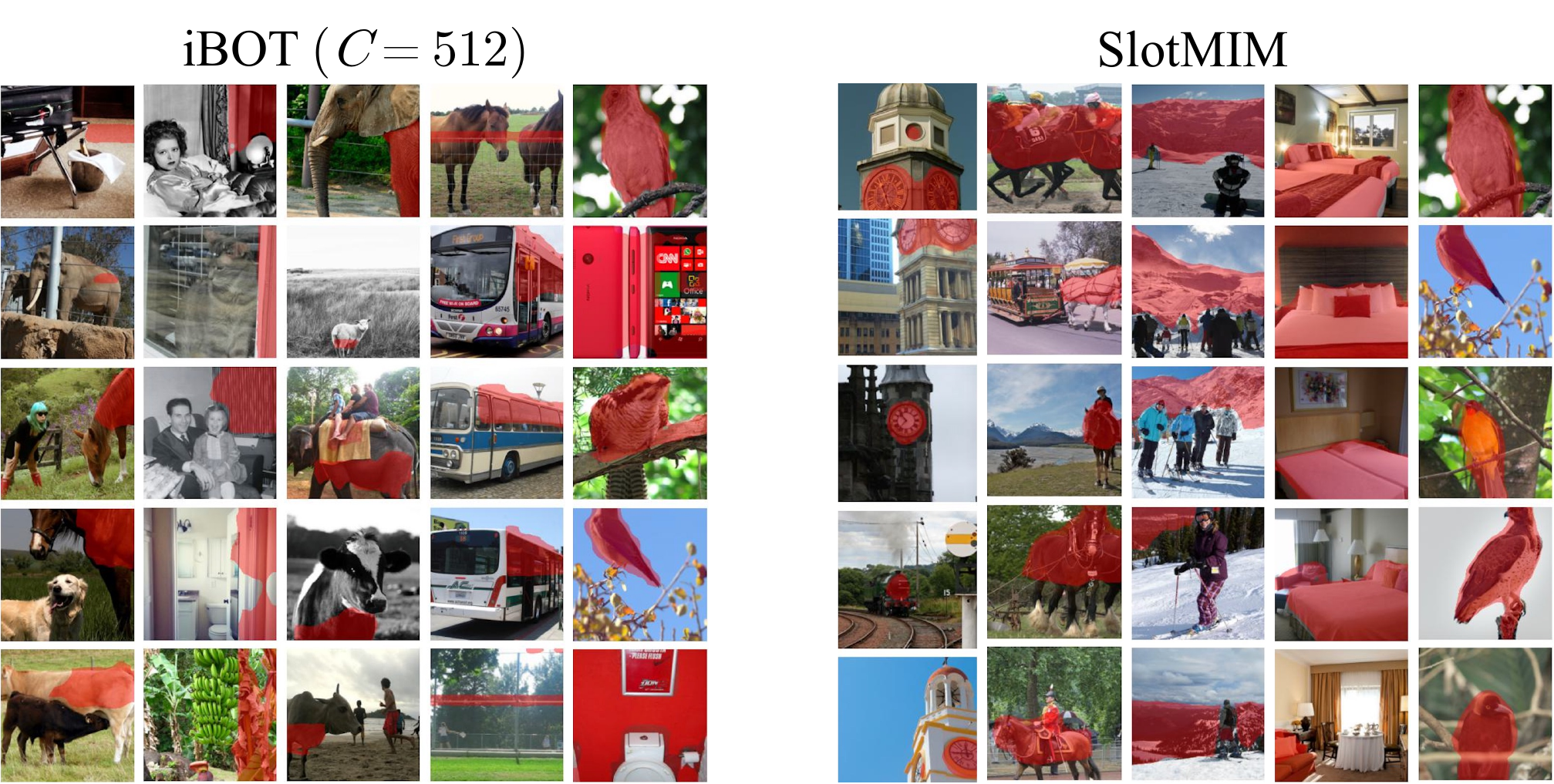}
        \caption{\textbf{Top-5 segments retrieved by the prototypes (by column).} A segment consists of patches assigned to the same prototype within an image. Each column shows the top-5 segments with the highest cosine similarity to the prototype corresponding to the column.}
        \label{fig:clusters_ibot_slotmim}
    \end{subfigure}
    \caption{
        \textbf{Comparison of concepts learned by iBOT and SlotMIM.}
        All models are trained on COCO+ for 800 epochs.
        While iBOT can discover fine-grained patterns, especially when using fewer prototypes (left), these patterns emerge bottom-up and lack semantic meaning. In contrast, SlotMIM's concepts are semantically coherent, making them more effective for instance discrimination pretext tasks (right).
        }
    \label{fig:compare_ibot_slotmim}
\end{figure*}

\paragraph{Visuomotor control tasks.}
To make the manipulation tasks better reflect the ability of PVMs, 1) we follow~\cite{vcond} to use trainable attentive pooling (see \cref{fig:bc_loop}), as opposed to prior works that employ \texttt{[CLS]} token. This is essential for a fair comparison between PVMs as the \texttt{[CLS]} token of some methods (\eg, MAE) does not receive any training signal and the results are rather arbitrary. Attentive pooling instead performs a learnable combination of all output tokens from the encoder, thus better utilizing the potential of models; 2) We also follow~\cite{hu2023pre} to avoid using any proprioceptive input to highlight the effect of PVMs, and 3) run 3 seeds for each experiment to avoid randomness. We then record the best performance of each run following~\cite{majumdar2023vc1}, and report average performance across seeds and tasks. We adopt 25 demonstrations per task. The benchmark covers 5 tasks in Franka Kitchen~\cite{gupta2019relay} and 8 tasks in Meta-World~\cite{yu2019meta}. Remaining details follows~\cite{hu2023pre}.
And all details in navigation tasks follow VC-1~\cite{majumdar2023vc1}.

\paragraph{Perception tasks.}
Following~\cite{burns2024what}, we report the segmentation Jaccard index of the best attention head on Pascal VOC 2012.
ADE20K semantic segmentation experiments follow the setting of MAE~\cite{he2021mae}, which uses UperNet~\cite{xiao2018upernet} and trains for 160K iterations with batch size 16.
For COCO object detection and instance segmentation, we follow the setting in iBOT~\cite{zhou2022ibot} to train a Cascade Mask R-CNN~\cite{cai2019cascade} with $1\times$ schedule (12 epochs), and report box and mask AP.

\subsection{Main Findings}

We evaluate models pre-trained on 241K-scale datasets, with complete results shown in \cref{fig:rl_voc_ade_241k}. For better understanding, we also visualize processed data in \cref{fig:teaser}\textcolor{cvprblue}{a} and \cref{fig:teaser}\textcolor{cvprblue}{b}. Below we discuss several key findings.

\paragraph{Neither MAE nor ego-centric data are optimal for visuomotor control.}
As shown in \cref{fig:rl_voc_ade_241k} (complete results) and \cref{fig:teaser}\textcolor{cvprblue}{a} (overall results), MAE and ego-centric data achieve only moderate performance in visuomotor control tasks, and perform poorly on ADE20K. This challenges the prevailing belief that MAE combined with ego-centric data is the best pre-training approach for PVMs, suggesting the need to explore alternative methods.

\paragraph{DINO/iBOT excel across tasks, especially with object-centric data.}
As demonstrated in \cref{fig:rl_voc_ade_241k} and \cref{fig:teaser}\textcolor{cvprblue}{a}, DINO and iBOT lead performance on most tasks, particularly when trained on object-centric data. iBOT also achieves strong semantic segmentation results when trained on scene-centric data. However, this performance advantage is sensitive to data distribution and doesn't generalize well across different pre-training datasets.

\paragraph{NOC data, particularly scene-centric data, significantly degrades top-performing methods.} While DINO/iBOT models achieve leading performance, they suffer substantial degradation on scene-centric data, as does MAE. This observation motivated us to investigate the underlying causes and develop a more robust PVM.

\paragraph{Objectness predicts DINO/iBOT performance.}
In \cref{fig:teaser}\textcolor{cvprblue}{b}, we demonstrate that DINO/iBOT's ability to learn objectness through attention maps deteriorates alongside performance drops on NOC data. The strong correlation (0.72) between objectness and task success rates aligns with findings from \cite{burns2024what}. This suggests that achieving good objectness from NOC data could lead to strong performance, which motivated our design of SlotMIM (introduced in \cref{sec:method}). As shown in \cref{fig:rl_voc_ade_241k}, this approach proves effective.

\section{Object-centric Learning on NOC Data}\label{sec:method}

\subsection{Preliminaries}

\paragraph{Deep clustering as self-distillation.}
DINO~\cite{caron2021dino} learns a set of $C$ prototypes online to cluster image embeddings. Given an input image $\vx \in \mathbb{R}^{H \times W \times 3}$, let $f_\theta$ and $f_\xi$ be student and teacher encoders that produce embeddings $\vz_\theta = f_\theta(\vx)$ and $\vz_\xi = f_\xi(\vx)$ respectively. The cluster assignments are computed as $p_\theta(\vx) = \text{softmax}(\vz_\theta \cdot \mathcal{C} / \tau)$, where $\mathcal{C} = \{\vc_c\}_{c=1}^C$ are the prototypes and $\tau$ is a temperature parameter. The loss is the cross-entropy between predictions of student and teacher models: $\mathcal{L}_\text{DINO}(\vv^1, \vv^2) = -\sum_{c=1}^C q_\xi(\vv^2)_c \log p_\theta(\vv^1)_c$, where $\vv^1$ and $\vv^2$ are two augmented views of the same image. Centering and sharpening operations are omitted in the equation for simplicity. Since it resembles knowledge distillation with soft labels produced by itself, DINO is also dubbed as self-distillation.

\paragraph{DINO on image patches with MIM.}
iBOT~\cite{zhou2022ibot} extends DINO to local image patches using masked image modeling (MIM). Given a binary mask $\mathcal{M} \in \{0,1\}^{N}$ indicating masked patches, the masked input $\tilde{\vv}$ replaces masked patches with a mask token $\vm$. The iBOT loss predicts cluster assignments of masked patches from unmasked ones: $\mathcal{L}_\text{iBOT}(\vv) = \sum_{i: \mathcal{M}_i = 1} \mathcal{L}_\text{DINO}(\tilde{\vv}_i, \vv_i)$, where $\tilde{\vv}_i$ is the masked patch from the student and $\vv_i$ is the corresponding unmasked patch from the teacher.

\paragraph{Slot attention}~\cite{locatello2020slot} is a variant of cross-attention that normalizes attention scores on the query axis instead of the key axis, encouraging queries to attend to different parts of the input. Our approach performs similar attentive pooling on patch embeddings based on their cluster assignments, with prototypes $\mathcal{C}$ acting as queries and patch embeddings $\vz_{\theta,i}$ as keys. Following convention, we refer to these pooled object features as slots -- prototypes adapted to image patches.

\subsection{SlotMIM Framework}

\paragraph{High-level intuition.}
We decompose self-supervised learning on NOC data into two subtasks: 1) learning to group image patches into objects, and 2) learning to discriminate objects as prior work done on object-centric data.
The key challenge is unsupervised object discovery, which we find emerges naturally from iBOT when using fewer prototypes.

\paragraph{Representation bottleneck for objectness.}
iBOT uses a set of prototype embeddings $\mathcal{C} = \{\vc_c\}_{c=1}^C$ to cluster image patches into $C$ groups, assigning each patch token a soft one-hot encoding $p_\theta(\vx_i)$ of its cluster membership. While iBOT typically uses $C=8192$ to capture fine patterns, we find a much smaller $C$ (e.g., 512 for COCO) better serves object discovery by creating an information bottleneck that encourages learning compositional object concepts. As shown in \cref{fig:attns_ibot_slotmim}, iBOT's clusters are very fine-grained (2nd row), but objectness emerges with fewer prototypes (3rd row).
However, these patterns still lack semantic meaning and can fragment single objects. Additionally, matching discovered objects between views remains difficult as their semantics vary despite being assigned to the same prototype (\cref{fig:clusters_ibot_slotmim}, left). Both issues suggest the need for semantic-level prototypes.

\paragraph{Cross-view consistency for semantic learning.}
The lack of semantic meaning stems from $\mathcal{L}_\text{iBOT}$ being computed between patches within the same view, providing no explicit guidance for learning view-invariant representations. We address this with a simple cross-view consistency objective $\mathcal{L}_\text{patch}^\text{cross}$ that encourages patches under different transformations to share the same token. We match patches between views using ROIAlign to align overlapping regions. Formally, for two augmented views $\vv^1$ and $\vv^2$ with patch embeddings $\tilde{\vz}_{\theta,i}^1 = f_\theta(\tilde{\vv}_i^1)$ and $\vz_{\xi,j}^2 = f_\xi(\vv_j^2)$, the loss is:
\begin{equation}
    \mathcal{L}_\text{patch}^\text{cross}(\vv^1, \vv^2) = -\frac{1}{|\mathcal{P}|} \sum_{(i,j) \in \mathcal{P}} \sum_{c=1}^C \vq_{\xi,i,c}^2 \log \tilde{\vp}_{\theta,j,c}^1 \,,
\end{equation}
where $\tilde{\vp}^1_\theta = \text{softmax}(\tilde{\vz}_\theta^1 \cdot \mathcal{C}_\theta / \tau_s)$ and $\vq^2_\xi = \text{softmax}(\vz_\xi^2 \cdot \mathcal{C}_\xi / \tau_t)$ are cluster assignments, $\tau_s$ and $\tau_t$ are temperature parameters, and $\mathcal{P}$ contains matched patch pairs.

\begin{figure*}[ht]
    \centering
    \includegraphics[width=\linewidth]{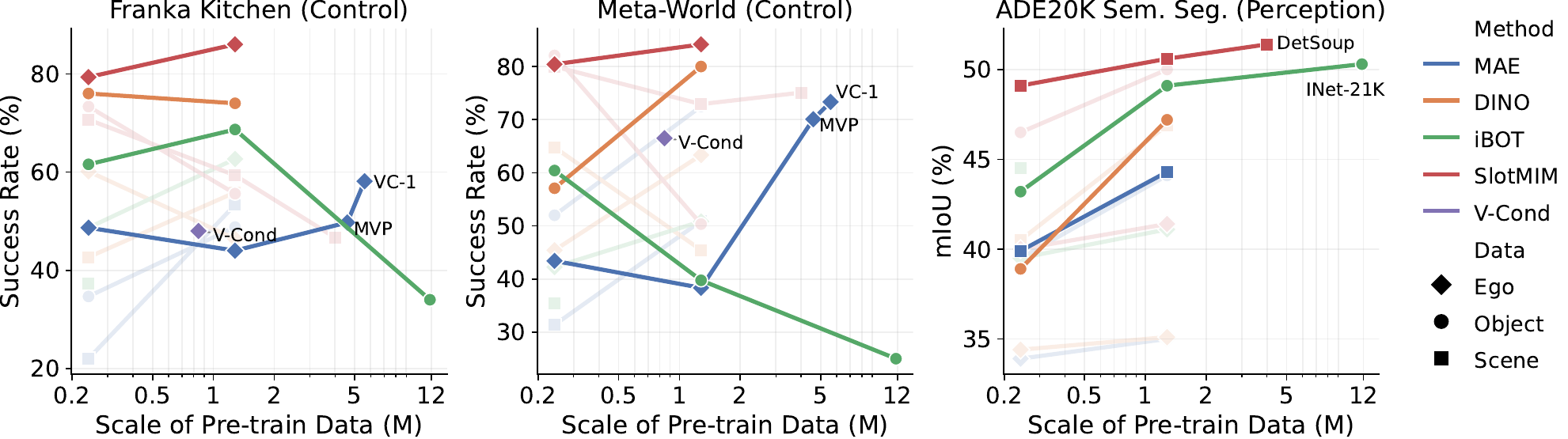}
    \caption{\textbf{Results of scaling PVM training data.} It considers three factors that influence manipulation success rates: data types, pre-training methods, and data scale. Highlighted lines represent the best-performing data scaling for each method, while \textcolor{gray}{faded} lines indicate sub-optimal performance. It shows that 1) SlotMIM achieves the best performance, scalability, and data efficiency across evaluation tasks by training on NOC data; 2) On manipulation tasks, most methods (except MAE) face performance drop when scaling up pre-training data.
    }
    \vspace{-1em}
    \label{fig:rl_ade_scale}
\end{figure*}

\paragraph{Object-level contrastive learning.}
With aligned object features, we apply contrastive learning at the object level. We only use slots that occupy at least one patch, filtering with the indicator: $\mathds{1}_i = \exists_j \text{ such that } \text{argmax}_c(p_\theta(\vv_j^1)_c) = i$. Slots with matching tokens form positive pairs. Using a MoCo-style approach with slots $\boldsymbol{s}_{\theta,i}^1 = h_\theta(\sum_j p_\theta(\vv_j^1)_i \vz_{\theta,j}^1)$, $\boldsymbol{s}_{\xi,i}^2 = \sum_j q_\xi(\vv_j^2)_i \vz_{\xi,j}^2$, the loss is:
\begin{equation}
    \mathcal{L}_\text{slot}(\tilde{\boldsymbol{s}}_\theta^1, \boldsymbol{s}_\xi^2) = -\frac{1}{K} \sum_{i=1}^C \log \frac{\mathds{1}_i^1 \mathds{1}_i^2 \exp(\boldsymbol{s}_{\theta,i}^1 \cdot \boldsymbol{s}_{\xi,i}^2 / \tau)}{\sum_{j=1}^C \mathds{1}_i^1 \mathds{1}_j^2 \exp(\boldsymbol{s}_{\theta,i}^1 \cdot \boldsymbol{s}_{\xi,j}^2 / \tau)} \,,
\end{equation}
where $h_\theta$ is a predictor MLP, $K=\sum_i \mathds{1}_i^1\mathds{1}_i^2$ counts positive pairs, and $\tau=0.2$. The final loss combines these objectives:
$\mathcal{L}_{\theta, \xi}(\tilde{\vv}^1, \vv^2) = \lambda_1 \mathcal{L}_\text{patch}^\text{within}(\tilde{\vv}^1, \vv^2) + \lambda_1 \mathcal{L}_\text{patch}^\text{cross}(\tilde{\vv}^1, \vv^2) + \lambda_2 \mathcal{L}_\text{slot}(\tilde{\boldsymbol{s}}_\theta^1, \boldsymbol{s}_\xi^2)$, in which $\mathcal{L}_\text{patch}^\text{within}$ is identical to $\mathcal{L}_\text{iBOT}$ and $\lambda_1=0.5$, $\lambda_2=1$. In practice, we optimize the symmetrized objective $\mathcal{L}_{\theta, \xi}(\tilde{\vv}^1, \vv^2) + \mathcal{L}_{\theta, \xi}(\tilde{\vv}^2, \vv^1)$.

\subsection{Ablation Study}
\begin{table}[ht]
\centering
\tablestyle{4pt}{1.05}
\vspace{-1em}
\begin{tabular}{lcccccccc}
 & mask & $\mathcal{L}_\text{patch}^\text{cross}$ & $\mathcal{L}_\text{patch}^\text{within}$ & $\mathcal{L}_\text{slot}$ & $k$-NN & ADE & Jacc & \textcolor{gray}{$\overline{K}$} \vspace{.5pt} \\
\toprule
1 & \colorbox{cyan!15}{\xmark} & \cmark & \colorbox{cyan!15}{\xmark} & \colorbox{cyan!15}{\xmark} & 45.1 & 47.4 & 42.5 & \textcolor{gray}{8.3} \\
2 & \cmark & \cmark & \colorbox{cyan!15}{\xmark} & \colorbox{cyan!15}{\xmark} & 44.9 & 48.6 & 42.3 & \textcolor{gray}{10.3} \\
3 & \cmark & \colorbox{cyan!15}{\xmark} & \cmark & \colorbox{cyan!15}{\xmark} & 27.7 & 45.7 & 39.3 & \textcolor{gray}{20.7} \\
4 & \cmark & \cmark & \colorbox{cyan!15}{\xmark} & \cmark & 45.3 & 47.5 & 42.9 & \textcolor{gray}{8.4} \\
5 & \cmark & \cmark & \cmark & \cmark & \textbf{46.2} & \textbf{49.1} & \textbf{43.9} & \textcolor{gray}{9.4} \\
\bottomrule
\end{tabular}%
\vspace{-.5em}
\caption{\textbf{Ablation study on effective modules.}}\label{tab:abl_module}
\vspace{-1em}
\end{table}

\noindent We ablate on several key components of SlotMIM. All models are trained on COCO+ for 800 epochs. As shown in \cref{tab:abl_module}, we evaluate each model variant using three metrics: $k$-NN accuracy on ImageNet classification, mean IoU on ADE20K semantic segmentation, and Jaccard index on VOC2012. We also report $\overline{K}$, the average number of objects/stuff regions discovered per image. The results reveal that Jaccard index positively correlates with representation quality, indicating that better objectness leads to stronger representations. Adding MIM improves downstream segmentation performance (comparing rows 1 and 2). The cross-view consistency loss and slot contrastive loss both contribute significantly to objectness (rows 2, 3, 5). Additionally, the within-view loss acts as an effective regularizer that further enhances the learned representations (rows 4, 5). Additional ablation study are available in the appendix.

\section{Scaling Up Pre-Training Data}

To provide an even more comprehensive picture, we extend \cref{fig:rl_voc_ade_241k} by scaling up the pre-training data from 241K to 1.28M and beyond (4M for SlotMIM, 5M for MAE, and 12M for iBOT).
The models at 1.28M scale follow the same settings as in \cref{tab:datasets}.
For SlotMIM, we combine ImageNet~\cite{deng2009imagenet}, COCO+~\cite{lin2014microsoft}, OpenImages~\cite{kuznetsova2020open}, Objects365~\cite{shao2019objects365}, and LVIS~\cite{gupta2019lvis} to create a 4M-scale scene-centric dataset, which we call DetSoup. We also utilize publicly available checkpoints of MAE trained on 5M-scale ego-centric datasets (MVP~\cite{radosavovic2023mvp} and VC-1~\cite{majumdar2023vc1}), and iBOT trained on 12M-scale ImageNet-21K~\cite{russakovsky2015imagenet}.
The results are presented in \cref{fig:rl_ade_scale}.

\subsection{Experiments on Manipulation Tasks}

The MAE regime includes MVP~\cite{radosavovic2023mvp} and VC-1~\cite{majumdar2023vc1}, which leverage MAE~\cite{he2021mae} to pre-train models on a massive collection of ego-centric videos~\cite{grauman2022ego4d} and Internet data. V-Cond~\cite{vcond} further proposes language-driven representation learning from human videos and their associated captions.
\cref{fig:rl_ade_scale} examines the relationship between manipulation success rates and pre-training methods, comparing scaling trends across different data types: ego-centric, object-centric, and scene-centric. Notably, increasing dataset size does not always improve performance across benchmarks, as also observed by~\cite{dasari2023an}.

\paragraph{What leads to inverse-scaling behaviors on manipulation tasks?}
In object manipulation tasks, scaling scene-centric and object-centric data to the million level can lead to performance drops for methods like DINO and iBOT---with MAE being the only exception. We hypothesize that self-supervised representation learning, including MIM, aims to learn invariance by pulling similar visual content together in the embedding space, effectively compressing visual data. However, scaling up data may result in over-compression, losing crucial low-level visual information necessary for visuomotor control tasks (\eg, accurate object grasping). MAE, on the other hand, continues to preserve low-level information due to the nature of its MIM objectives. This also explains why MAE is commonly preferred in existing PVM works---they typically start with million-scale data, and MAE is one of the few methods that avoid over-compression at such scale.

\paragraph{Why SlotMIM does not face over-compression when trained on ego-centric data?}
SlotMIM builds its training objective on concepts it discovers, with the type of discovered concepts determined by the training data distribution. As shown in \cref{fig:teaser}\textcolor{cvprblue}{d}, SlotMIM learns coarse-grained objects from object/scene-centric data, but fine-grained parts from ego-centric data, achieving a good balance between invariance and objectness. We believe this occurs because, unlike object/scene-centric data from diverse Internet sources, ego-centric images come from consecutive human videos sharing contextual backgrounds or scenarios. This contextual similarity means invariance learning in ego-centric data focuses more on differences within the same video or scenario, particularly in foreground objects. This focus is crucial for robot manipulation learning, which requires effective interaction with these foreground objects.

\paragraph{SlotMIM is more data efficient and scalable in leveraging ego-centric data.}
Compared to general-purpose pre-trained models and state-of-the-art robot learning methods (\eg, MVP~\cite{radosavovic2023mvp} and VC-1~\cite{majumdar2023vc1}), we demonstrate that SlotMIM, pre-trained with just 241K data samples, outperforms prior methods that used over 1M samples. When scaled to 1M ego-centric data, it achieves the highest success rates among all methods in the comparison.

\subsection{Experiments on ADE20K and COCO}
As shown in \cref{fig:rl_ade_scale} (right), compared to previous efforts scaling up with ImageNet-21K (12M images)~\cite{russakovsky2015imagenet}, our SlotMIM models continue to improve and surpass them using 3$\times$ less data. This suggests that NOC data can be a more scalable learning resource.

\begin{figure}[ht]
    \centering
    \includegraphics[width=\linewidth]{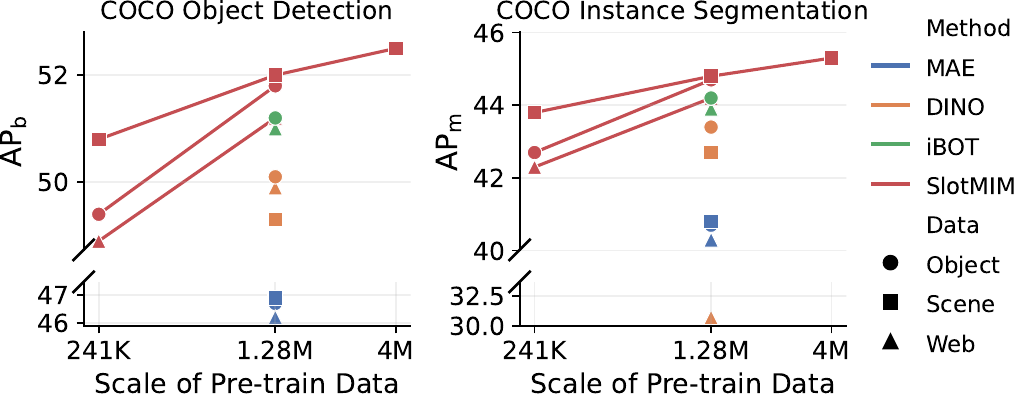}
    \caption{\textbf{Results on COCO object detection and instance segmentation.} SlotMIM shows better data efficiency with both object-centric and NOC data, and its results continue to improve with more data, surpassing prior models by a notable margin.}\label{fig:det}
\end{figure}
    
In \cref{fig:det}, we present an evaluation on COCO object detection and instance segmentation. The superiority of SlotMIM is evident and continues to improve with increased data scale. As supported by the visualizations in \cref{fig:teaser}\textcolor{cvprblue}{d}, SlotMIM learns to discover common objects from scenes in alignment with human visual perception, supporting its scaling capability for segmentation/detection tasks.

\subsection{Experiments on Navigation Tasks}

\begin{table}[ht]
    \centering
    \tablestyle{2pt}{1.05}
    \begin{tabular}{cccccc}
    Method  & Arch     & Pretrain Data & Scale & ObjectNav     & ImageNav      \\
    \toprule
    MVP~\cite{radosavovic2023mvp}     & ViT-B & EgoSoup       & 4.6M  & 51.2          & 64.7          \\
    VC-1~\cite{majumdar2023vc1}    & ViT-B & Ego4D+MNI     & 5.6M  & 55.4          & 67.9          \\
    SlotMIM & ViT-B & Ego4D         & 1.28M & 48.4          & 65.4          \\
    SlotMIM & ViT-B & DetSoup    & 4.0M    & \textbf{62.0} & \textbf{69.8}          \\
    \midrule
    \textcolor{gray}{MVP~\cite{radosavovic2023mvp}}     & \textcolor{gray}{ViT-L} & \textcolor{gray}{EgoSoup}       & \textcolor{gray}{4.6M}  & \textcolor{gray}{55.0}          & \textcolor{gray}{68.1}          \\
    \textcolor{gray}{VC-1~\cite{majumdar2023vc1}}    & \textcolor{gray}{ViT-L} & \textcolor{gray}{Ego4D+MNI}     & \textcolor{gray}{5.6M}  & \textcolor{gray}{60.3}          & \textcolor{gray}{\textbf{70.3}} \\
    \bottomrule      
    \end{tabular}
    \caption{\textbf{Results on ObjectNav and ImageNav.} Compared with prior SoTA methods using ViT-B, SlotMIM trained on DetSoup improves by 6.6\% and 1.9\% on ObjectNav and ImageNav, respectively. Compared with ViT-L models, it still rivals them on ObjectNav and achieves comparable performance on ImageNav.}\label{tab:nav}
\end{table}

\noindent In \cref{tab:nav}, we compare SlotMIM's performance on ObjectNav and ImageNav benchmarks with SoTA methods. Details of the navigation tasks can be found in \cref{tab:robo_tasks}. Due to the extreme GPU demands of these tasks, we evaluated only two SlotMIM models: each trained on the largest-scale scene/ego-centric data (DetSoup and Ego4D) available. SlotMIM trained on Ego4D shows moderate performance on ImageNav but underperforms on ObjectNav, indicating a misalignment with downstream tasks. In contrast, when trained on DetSoup, it outperforms MVP and VC-1 by 6.6\% and 1.9\% on ObjectNav and ImageNav, respectively. This demonstrates both the effectiveness of SlotMIM and suggests navigation are closely related to perception tasks.

\section{Conclusion} %
This work revisits the potential of PVMs for robot learning tasks.
We construct a comprehensive benchmark that evaluates models on robot learning tasks to find out what is the best pretraining method and what is the best data for pretraining.
Our findings suggest that while DINO and iBOT lead the benchmark, their performance degrades rapidly when trained on NOC data.
Furthermore, this degradation can be partially alleviated by strengthening the ability to learn object-centric representations.
Motivated by these findings, we propose SlotMIM to effectively learn object-centric representations from NOC datasets.
Through extensive experiments across diverse datasets and downstream tasks, including robotics, we demonstrated the consistent superiority of SlotMIM over existing methods.
We hope our promising results open new avenues for adopting PVMs for more robot learning tasks.

\section*{Acknowledgements}
This work has been supported in part by Hong Kong Research Grant Council --- Early Career Scheme (Grant No. 27209621), General Research Fund Scheme (Grant No. 17202422, 17212923), Theme-based Research (Grant No. T45-701/22-R) and Shenzhen Science and Technology Innovation Commission (SGDX20220530111405040).  Part of the described research work is conducted in the JC STEM Lab of Robotics for Soft Materials funded by The Hong Kong Jockey Club Charities Trust. 

{
    \small
    \bibliographystyle{ieeenat_fullname}
    \bibliography{main}
}

\addtocontents{toc}{\protect\setcounter{tocdepth}{2}}
\appendix
\clearpage
\maketitlesupplementary

{
  \hypersetup{linkcolor=black}
  \tableofcontents
}

\section{Extended Implementation Details}

\subsection{Pre-Training Details}

\paragraph{Architecture.} We use ViT-B/16~\cite{dosovitskiy2021vit} as our backbone. Following common practice in DINO~\cite{caron2021dino}, iBOT~\cite{zhou2022ibot} and MoCo-v3~\cite{chen2021mocov3}, the projector $g$ is a $3$-layer MLPs with hidden dimension $2048$ and output dimension $256$, and the predictor $h$ is a $2$-layer MLPs with hidden dimension $4096$ and output dimension $256$.

\paragraph{Augmentation and masking.} We use the same augmentation strategy as in iBOT~\cite{zhou2022ibot} except not using small local crops (multi-crop). Avoiding the use of multi-crop saves significant computational costs in our model, and the model-learned slots work in a similar role. The masking strategy follows iBOT~\cite{zhou2022ibot}, with prediction ratio $r$ uniformly sampled from range $[0.3-0.2, 0.3+0.2]$.

\paragraph{Optimization.} Most optimization configurations follow DINO~\cite{caron2021dino} and iBOT~\cite{zhou2022ibot}. We use AdamW optimizer with a cosine schedule for the learning rate and weight decay. The learning rate is linearly ramped up during the first $10$ epochs to $1.5\times10^{-4}$ scaled with the total batch size: $lr = lr_\text{base} \times \text{batch size}/256$, and then decays following the cosine schedule. The weight decay starts from $0.4$ and also decays following the cosine schedule, to $0.04$ when training ends.
We train for $800$ epochs on 241K-scale datasets and $400$ epochs on 1.28M-scale datasets, with a batch size of $1024$ distributed across 8 A100 GPUs. For experiments on 4M-scale datasets, we train $200$ epochs.

\paragraph{Hyperparameters.} Follow DINO~\cite{caron2021dino} and iBOT~\cite{zhou2022ibot}, the teacher temperature $\tau_t$ linearly ramps up from $0.04$ to $0.07$ for the first $30$ epochs and remains constant afterwards. The student temperature $\tau_s$ is fixed at $0.1$. The number of prototypes $C$ is set to $512$ for COCO+ and $1024$ for other datasets. 

\subsection{Evaluation Details}

\paragraph{Manipulation tasks.} Following the setup of~\cite{hu2023pre}, we use a shallow 4-layer MLP with hidden sizes $[512, 256, 128]$ and ReLU activations as the policy network for behavior cloning. The 5 Franka Kitchen tasks, 8 Meta-World tasks, and corresponding GT demonstrations are also taken from~\cite{hu2023pre}. Following R3M~\cite{nairr3m} and VC-1~\cite{majumdar2023vc1}, the policy training involves mini-batches of $128$ samples, conducted over $20000$ steps with the Adam optimizer and a learning rate of $0.001$. The model is evaluated every $1000$ steps. All tasks and environments use $224\times224$ RGB images without proprioceptive input and without image augmentations, and each task uses only 25 demonstrations for training, which raises higher requirements on the quality of PVMs. For better measurement of the potential of PVMs, we follow~\cite{vcond} and use attentive pooling (also known as multihead attention pooling~\cite{lee2019set}), which is suggested as a strong and versatile approach~\cite{zhai2022scaling} as opposed to the commonly used \texttt{[CLS]} token and provides better comparisons between frozen PVMs.
Following VC-1~\cite{majumdar2023vc1}, we take the best checkpoint for each run on each task. For each environment, we report the average performance over all tasks ($3$ independent runs each task).

\paragraph{Navigation tasks.} The settings on navigation tasks strictly follows VC-1~\cite{majumdar2023vc1}, which involves object-goal navigation~\cite{batra2020objectnav} and image-goal navigation~\cite{zhu2017imagenav}. In both, the agent is initialized at a random location in an unknown 3D environment and is tasked to find the goal location specified by an image or object. Both tasks are conducted using Habitat~\cite{savva2019habitat} simulator, in which ObjectNav is conducted in the HM3D~\cite{ramakrishnan2021hm3d} environment and ImageNav is conducted in the Gibson~\cite{xia2018gibson} environment.
For ObjectNav, the agent is trained for approximately $400$M steps with $512$ parallel environments. For ImageNav, the agent is trained for $500$M steps with $320$ parallel environments. Further details can be found in~\cite{majumdar2023vc1} and omitted here for brevity.

\paragraph{Semantic segmentation on ADE20K.} We use UperNet~\cite{xiao2018upernet} implemented in MMSegmentation following iBOT~\cite{zhou2022ibot}. Specifically, we fine-tune for $160\mathrm{k}$ iterations with stochastic gradient descent, with a batch size of $16$ and weight decay of $0.0005$. The learning rate is $0.01$ and decays following the poly schedule with power of $0.9$ and min\_lr of $0.0001$.

\paragraph{Object detection and instance segmentation on COCO.}
COCO object detection and instance segmentation setting also follows iBOT~\cite{zhou2022ibot}, where the pre-trained model initialized a Cascade Mask R-CNN~\cite{cai2019cascade}.
The image scale is $[640, 800]$ during training and $800$ at inference. We fine-tune all layers end-to-end on COCO~\cite{lin2014microsoft} \texttt{train2017} set with the standard $1\times$ schedule and report AP for boxes and masks on the \texttt{val2017} set.

\paragraph{Analytical metrics.}
Some numeric indicators are considered to help analyze properties of pre-trained models, \eg, object discovery ability (objectness of attention maps) measured by Pascal VOC 2012 object segmentation quality. The Jaccard index measures the overlap between predicted mask $P$ and the ground truth mask $G$ as $J(P, G) = \frac{G\cap P}{G\cup P}$. Following~\cite{burns2024what,naseer2021intriguing}, the attention maps of DINO and iBOT are computed between the \texttt{[CLS]} token and patch tokens in the last layer, and the attention maps of SlotMIM are computed between the prototypes and projected patch tokens. For each object of interest, the attention head/prototype producing the best Jaccard index is selected.
Besides, for the ablation studies, we also report $k$-NN ImageNet classification ($k=20$) accuracy following DINO~\cite{caron2021dino}. Additionally, we maintain a running mean of the average number of active (assigned to at least one patch) slots $\overline{K}$ in an image during training.

\subsection{Details for Fig. \textcolor{cvprblue}{1}}

\paragraph{(a) BC performance regarding dataset and model.}
Fig.~\textcolor{cvprblue}{1a} is a grouped version of Franka Kitchen and Meta-World results in Fig.~\textcolor{cvprblue}{2} in the main paper. For each grid, we report the average performance over all manipulation tasks given a PVM pre-trained on a specific dataset (by row) using a specific model (by column).

\paragraph{(b) Correlation between objectness and BC performance.}
Fig.~\textcolor{cvprblue}{1b} is a joined view between the average manipulation performance (first two subfigures) and VOC object segmentation performance (third subfigure) in Fig.~\textcolor{cvprblue}{2} in the main paper. It is presented as a scatter plot to show the (Pearson's) correlation between the two metrics.

\paragraph{(d) Visualization of attention maps.}
Fig.~\textcolor{cvprblue}{1b} and Fig.~\textcolor{cvprblue}{1d} show examples of the attention maps of DINO and SlotMIM, respectively. All models are pre-trained on 241K-scale datasets for 800 epochs. For DINO, we follow the official implementation~\cite{caron2021dino} to take the attention maps between the \texttt{[CLS]} token and patch tokens in the last layer, and show the best attention head. For SlotMIM, we take the attention maps (prototype assignments) between projected patch tokens and the ``cat'' prototype---the prototype with the highest cosine similarity with ``cat'' segments.

\subsection{Details for Fig.~\textcolor{cvprblue}{4}}

\paragraph{(a) Visualization of attention maps.}
All attention maps in Fig.~\textcolor{cvprblue}{4a} are computed in the same manner as SlotMIM in Fig.~\textcolor{cvprblue}{1d}---between the prototypes and projected patch tokens. All models are pre-trained on COCO+ (scene-centric) for 800 epochs. For iBOT, we trained two variants---one with 8192 prototypes (dimension of the last layer in the DINOHead in implementation) as default, and one with 512 prototypes. In the visualization, each prototype is assigned a random color.

\paragraph{(b) Visualization of segmentation consistency.}
Fig.~\textcolor{cvprblue}{4b} shows the segments assigned to each prototype following the implementation of~\cite{wen2022slotcon}. We first obtain all segments (prototype assignments) on COCO \texttt{val2017} set, pool them to slots, and then retrieve the nearest-neighbor slots for each prototype. Then for each selected prototype, we visualize the segments of the top-5 similar slots.

\begin{table*}[ht]
    \centering
    \subfloat[
    \textbf{Number of prototypes} \label{subtab:slotnum}
    ]{%
    \begin{minipage}[t]{0.24\linewidth}
        \centering
        \setlength{\tabcolsep}{2pt}\renewcommand{\arraystretch}{1.15}
        \small
        \begin{tabular}[t]{lcccccc}
        $C$ & $k$-NN & ADE & Jacc & \textcolor{gray}{$\overline{K}$} \\ 
        \toprule
        256  & 45.3 & \textbf{49.1} & 42.2 & \textcolor{gray}{7.8} \\
        \rowcolor{cyan!10}
        512  & \textbf{46.2} & \textbf{49.1} & \textbf{43.9} & \textcolor{gray}{9.4} \\
        1024 & 45.6 & 48.4 & 42.8 & \textcolor{gray}{10.8} \\
        \end{tabular}
    \end{minipage}
    }
    \subfloat[
    \textbf{Mask ratio ($\pm$0.2)} \label{subtab:maskratio}
    ]{%
    \begin{minipage}[t]{0.22\linewidth}
        \centering
        \setlength{\tabcolsep}{2pt}\renewcommand{\arraystretch}{1.15}
        \small
        \begin{tabular}[t]{lcccccc}
            & $k$-NN & ADE & Jacc & \textcolor{gray}{$\overline{K}$} \\ 
            \toprule
            \rowcolor{cyan!10}
            0.3 & \textbf{46.2} & \textbf{49.1} & 43.9 & \textcolor{gray}{9.4} \\
            0.4 & 45.8 & 48.6 & 45.0 & \textcolor{gray}{8.1} \\
            0.5 & 44.3 & 48.2 & \textbf{45.7} & \textcolor{gray}{7.1} \\
            \end{tabular}
    \end{minipage}
    }
    \subfloat[
    \textbf{Teacer temp. schedule} \label{subtab:temp}
    ]{%
    \begin{minipage}[t]{0.27\linewidth}
        \centering
        \setlength{\tabcolsep}{2pt}\renewcommand{\arraystretch}{1.15}
        \small
        \begin{tabular}[t]{lcccccc}
            $\tau_t$ & $k$-NN & ADE & Jacc & \textcolor{gray}{$\overline{K}$} \\ 
            \toprule
            \rowcolor{cyan!10}
            0.04$\rightarrow$0.07 & \textbf{46.2} & \textbf{49.1} & \textbf{43.9} & \textcolor{gray}{9.4} \\
            0.07 & 45.8 & 48.6 & 42.1 & \textcolor{gray}{8.1} \\
            0.07$\rightarrow$0.04 & 45.5 & \textbf{49.1} & 42.6 & \textcolor{gray}{7.1} \\
            \end{tabular}
    \end{minipage}
    }
    \subfloat[
    \textbf{Patch loss} \label{subtab:loss}
    ]{%
    \begin{minipage}[t]{0.25\linewidth}
        \centering
        \setlength{\tabcolsep}{2pt}\renewcommand{\arraystretch}{1.15}
        \small
        \begin{tabular}[t]{lcccccc}
            Type & $k$-NN & ADE & Jacc & \textcolor{gray}{$\overline{K}$} \\ 
            \toprule
            \rowcolor{cyan!10}
            center & \textbf{46.2} & 49.1 & \textbf{43.9} & \textcolor{gray}{9.4} \\
            SH & 45.1 & \textbf{49.3} & 40.8 & \textcolor{gray}{15.2} \\
            \\
        \end{tabular}
    \end{minipage}
    }
    \caption{\textbf{Ablation studies on hyperparameters.} Default values are marked with a \colorbox{cyan!10}{cyan} background.} 
    \label{tab:ablation:hyper}
\end{table*}

\section{Extended Analysis and Discussion}

\subsection{Why Pre-Training on NOC Data?}
Self-supervised pre-training have benefited numerous downstream tasks, with a key advantage in its ability to learn representations from unlabeled data, eliminating the need for human annotations and making it easier to scale up training datasets. Despite this advantage in utilizing diverse types of data, most research has focused on (single-)object-centric datasets like ImageNet for model development, leaving large volumes of non-object-centric (NOC) data, such as Open Images~\citep{kuznetsova2020open}, SA-1B~\citep{kirillov2023segment}, LAION~\citep{schuhmann2022laion5b}, and Ego4D~\citep{grauman2022ego4d}, underutilized. However, many primary application domains of self-supervised learning---such as robot learning (manipulation, navigation, \etc), or traditional perception tasks like object detection and image segmentation---often require handling NOC data. 
This motivates us to explore the potential of NOC data for self-supervised learning, which could bridge the data-domain gap between self-supervised learning and real-world applications, and is rich in information, offering new opportunities for data scaling.

\subsection{Why Consider Both Control and Perception?}
The latest development of robot learning has witnessed a shift towards mulit-modal generalist models that are capable of handling diverse robot tasks, in which both control and perception are indispensable. This involves integrating PVMs (\eg, CLIP~\cite{radford2021clip}, T5~\cite{raffel2020t5}, and DINOv2~\cite{oquab2023dinov2}) as multi-modal encoders~\cite{kim2024openvla,octo2024,rt12023,rt22023,wu2024gr1,cheang2024gr2} or explicitly utilizing foundation models (\eg, OWL-ViT~\cite{minderer2022simple}, SAM~\cite{kirillov2023segment}, and GPT-4V~\cite{openai2023gpt4}) as tools~\cite{hu2023vila,huang2023voxposer,huang2024copa,fangandliu2024moka,pivot2024}. It is also shown that improvements in perception can lead to policies that generalize better to unseen environments using less data~\cite{moo2023}.
In accordance with this trend, recent research in PVM for robot learning has also started to consider more diverse evaluation protocols~\cite{majumdar2023vc1,vcond,zeng2024mpi}. Serving as the visual cortex of a modern robot, we believe that a good PVM should enhance both perception and control abilities.

\subsection{What Determines the Objectness of SlotMIM?}
We consider two quantitative metrics for the objectness of SlotMIM: 1) the Jaccard index between the attention maps and the ground truth object masks in Pascal VOC; and 2) the average number of active slots $\overline{K}$ in an image during training.
The former measures the \textit{alignment} of the attention maps to objects, and the latter roughly measures the \textit{granularity} of the concepts represented by the attention maps---the more fine-grained the concepts are, the more parts (slots) each image is segmented into.

In the main paper, we have shown that both pre-training dataset and model hyper-parameters can affect the objectness of SlotMIM. Concerning the dataset, object-centric data leads to better alignment to common objects, scene-centric data and web-crawled data lead to slightly worse segmentation quality and similar-level granularity, and ego-centric data leads to very fine-grained segmentation. Concerning the model hyper-parameters, we will show in \cref{subsec:ablation_ext} that there are multiple factors can affect the segmentation quality and granularity.

\subsection{Fine-grained Slots for Manipulation Tasks?}
In the main paper, we have shown that while scaling up non-ego-centric data for SlotMIM can improve segmentation quality and navigation/perception performance, manipulation performance drops due to over-compression. We are curious if learning fine-grained slots/concepts can improve the transferability of SlotMIM pre-trained on non-ego-centric data to manipulation tasks. To this end, we consider pre-training on COCO+ and conduct an ablation study on the number of prototypes and the use of Sinkhorn-Knopp algorithm for patch-level loss. The results are shown in \cref{tab:fine_grain_bc}.

\begin{table}[ht]
\centering
\tablestyle{4pt}{1.05}
\begin{tabular}{ccccccc}
    Model & Dataset & \#Proto. & SH & Jacc & $\overline{K}$ & Success (\%) \\
    \toprule
    SlotMIM & COCO+ & 256 & \xmark & 42.2 & 7.8 & 74.3 \\
    SlotMIM & COCO+ & 512 & \xmark & \textbf{43.9} & 9.4 & 74.8 \\
    SlotMIM & COCO+ & 512 & \cmark & 40.8 & 15.2 & 76.8 \\
    SlotMIM & COCO+ & 1024 & \xmark & 42.8 & 10.8 & \textbf{78.4} \\
    \toprule
\end{tabular}
\caption{\textbf{Can fine-grained slots improve manipulation performance?} We consider increasing the number of prototypes $C$ and using SH for patch-level loss. Both ways enforce the emergence of fine-grained slots and improve success rate on manipulation tasks.}\label{tab:fine_grain_bc}
\end{table}

Increasing the number of prototypes $C$ loosens the constraints on the compactness of the slots, and the use of SH encourages more diverse utilization of the prototypes, both leading to more fine-grained slots. The results show that both methods can improve the success rate on manipulation tasks (averaged over all tasks).
Explorations in this direction may resolve the inverse-scaling phenomenon on non-ego-centric data in the main paper, which we leave for future work.

\subsection{Limitation and Future Work}
This work explores the interaction between pre-training data and algorithms for robotic manipulation, navigation, and perception-oriented tasks. With these of interest, it requires extremely intensive computation to provide a complete comparison of all existing PVMs, scale pre-training data to larger scales, and evaluate more complex robotic tasks (\eg, language-conditioned manipulation and real-world tasks). The results present in the paper is thus a result of trading-off. Will SlotMIM work well for generalist robotic models like Octo~\cite{octo2024} and OpenVLA~\cite{kim2024openvla}? Will the next-gen PVM for robotics be pre-trained on a mixture of SA-1B~\cite{kirillov2023segment}, Ego4D~\cite{grauman2022ego4d}, and/or LAION-400M~\cite{schuhmann2021laion400m}, LVD-142M~\cite{oquab2023dinov2}? Will the preceptive module bestly to be supervised by self-supervision, language, action trajectories, or a mixture of them?
There are yet much to explore, but still, we believe that the insights and methods proposed in this work can serve as a stepping stone for future research in this direction.

\section{Extended Experiments}

\subsection{Extended Ablation Study} \label{subsec:ablation_ext}
In \cref{tab:ablation:hyper} we present ablations on some numeric design choices. Generally speaking, a smaller number of prototypes, a higher mask ratio, and the use of centering~\cite{caron2021dino} instead of Sinkhorn-Knopp algorithm~\cite{caron2020swav} encourage the network to discover more holistic concepts/objects, while the opposite discovers more fine-grained ones. Optimal representation is highly related to object discovery quality.

\subsection{Comparison with DINOv2}
\begin{table}[!h]
    \centering
    \tablestyle{2.1pt}{1.05}
    \begin{tabular}{lcccccc}
        Method & Dataset & Scale & Kitchen & MW & ObjNav & ImgNav  \\
        \toprule
        MVP & EgoSoup & 4.6M & 49.7 & 70.1 & 51.2 & 64.7 \\
        VC-1 & Ego4D+MNI & 5.6M & 58.1 & 73.3 & 55.4 & 67.9 \\
        DINOv2 & LVD & 142M & 64.0 & 38.8 & \textbf{65.8} & 59.1 \\
        SlotMIM & Ego4D & 1.28M & \textbf{86.0} & \textbf{84.2} & 48.4 & 65.4 \\
        SlotMIM & DetSoup & 4M & 46.7 & 75.1 & 62.0 & \textbf{69.8} \\
        \toprule
    \end{tabular}
    \caption{\textbf{Comparison with DINOv2.} SlotMIM achieves comparable or better performance on robot tasks compared to DINOv2, especially on manipulation tasks. (DINOv2 is ViT-B/14, while other models are ViT-B/16)}\label{tab:dino2}
\end{table}

One might argue that the state-of-the-art self-supervised model, DINOv2~\citep{oquab2023dinov2}, already utilizes NOC data with a vision transformer backbone. However, its success heavily depends on data curation techniques that leverage the object-centric ImageNet dataset to select neighboring data from web-crawled data, keeping its data distribution closely tied to object-centric approaches. We also evaluate DINOv2 on the robot learning tasks considered in this paper. As shown in \cref{tab:dino2}, DINOv2 does not perform as well on these tasks, possibly also due to over-compression of the representations for manipulation tasks.

\subsection{ImageNet Linear Probing and Fine-tuning} \label{subsec:baseline_compute}
\paragraph{Setting.} We follow MAE~\cite{he2021mae} for details on ImageNet evaluations. For linear probing, we insert an extra BatchNorm layer without affine transformation between the features and the linear classifier. We train with batch size 4096, initial learning rate 0.1, and optimize using SGD for 90 epochs. We sweep between \texttt{[CLS]} token and average pooling and report the best results of pre-trained models.
For fine-tuning, we train a linear classifier on frozen features for 100 epochs using SGD with momentum 0.9, batch size 1024, and initial learning rate 1e-3 with cosine decay. We follow MAE~\cite{he2021mae} to adopt average pooling.
For both settings, accuracy is evaluated on a single 224$\times$224 crop.

\begin{figure}[ht]
    \centering
    \includegraphics[width=\linewidth]{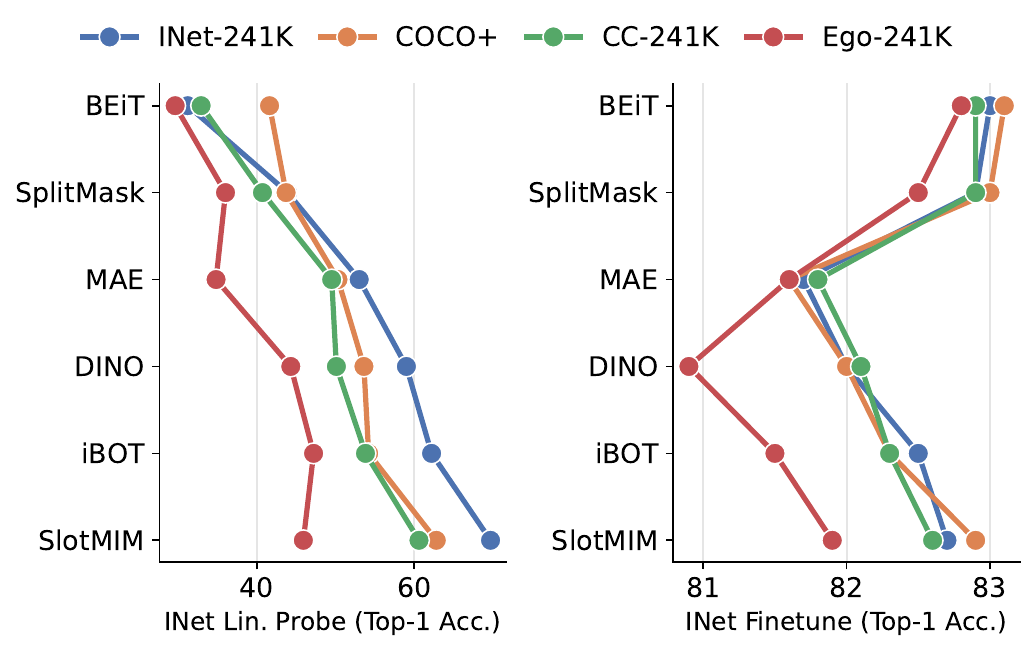}
    \caption{
    \textbf{Results on ImageNet tasks.}
    SlotMIM consistently outperforms prior arts whether pre-trained on object-centric data or not. Notably, when trained on COCO+, it transfers better than most ImageNet models despite the domain gap.}
    \label{fig:241k_all}
\end{figure}

We first evaluate models pre-trained on 241K-scale datasets, and show that NOC data can be good learning resources if used properly. The results are present in \cref{fig:241k_all}. Overall, SlotMIM achieves the best performance across classification and segmentation tasks, no matter learning from object-centric data or not. Below, we discuss some other interesting findings.

\paragraph{Features learned from NOC data can be linear separatable on ImageNet.}
From \cref{fig:241k_all} (left), our models trained on COCO and CC achieve similarly good linear probing performance on ImageNet with best prior ImageNet-trained methods. As a clear contrast, all previous methods trained on NOC datasets (COCO, CC, and Ego4D) fall behind the best ImageNet counterpart.

\paragraph{NOC data can be worth more than ImageNet for ImageNet.}
As shown in \cref{fig:241k_all} (right), under ImageNet fine-tuning setting, the top-3 methods (BEiT, SplitMask, and SlotMIM) have the best performance when trained on COCO+ instead of ImageNet. For MAE and DINO, training on CC also transfers better than ImageNet. Note that this is uncommon given the domain gap between NOC pre-training data and OC downstream task, demonstrating that NOC data are information-rich learning resources.

\subsection{Scaling Up for ImageNet Tasks} \label{subsec:scale_up_data}
\begin{figure}[ht]
    \centering
    \includegraphics[width=\linewidth]{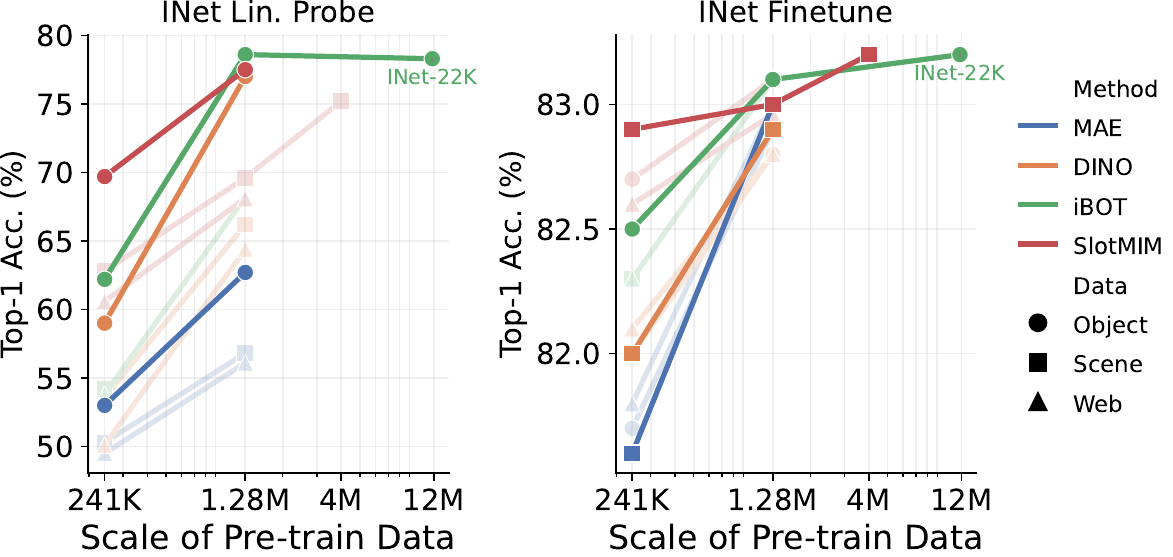}
    \caption{\textbf{Scaling on different data sources.} We scale up object-centric, scene-centric, and web-crawled data, and highlight the best (model, data) combinations. Our method learns strong and transferable representations with significant data efficiency and continues to improve with more data.}
    \label{fig:more_data}
\end{figure}

Superior data efficiency allows us to explore larger-scale pre-training data. In \cref{fig:more_data}, we show that SlotMIM achieves strong performance with remarkable data efficiency.

\paragraph{Comparable or better performance with small data scale.}
As shown in Figure \ref{fig:more_data}, SlotMIM achieves comparable or superior performance to other methods using significantly less data. Our INet-241K model for ImageNet linear probing, and COCO+/INet-241K models for ImageNet fine-tuning outperform or match most models trained on 1.28M ImageNet images across various tasks. This remarkable data efficiency demonstrates our approach's effectiveness in extracting rich, transferable features from limited data.

\paragraph{NOC pre-training rivals ImageNet pre-training for ImageNet.}
Interestingly, we observe that pre-training on NOC datasets like OpenImages-1.28M can lead to performance better than pre-training on ImageNet for the ImageNet classification task (fine-tuning setting). When scaled up to 4M scale, this trend becomes more pronounced. This aligns with the trend in \cref{fig:241k_all} that NOC data can provide more information-rich features, which can be better-utilized by models like SlotMIM.

\paragraph{NOC data also possesses stronger scalability.}
We extend experiments to 4M scale by combining ImageNet~\cite{deng2009imagenet}, COCO+~\cite{lin2014microsoft}, OpenImages~\cite{kuznetsova2020open}, Objects365~\cite{shao2019objects365}, and LVIS~\cite{gupta2019lvis}.
Compared with previous efforts on scaling up with ImageNet-22K~\cite{russakovsky2015imagenet} (12M images), the performance of SlotMIM models continues to grow and surpasses them with 3$\times$ less data.
This suggests that NOC data can be a more scalable learning resource.

\end{document}